\documentclass[10pt,twocolumn,letterpaper]{article}

\usepackage{cvpr}              

\usepackage[T1]{fontenc}

\usepackage{gradient-text}

\newcommand{\refedit}{\textbf{\gradientRGB{RefEdit}{0,76,153}{102,178,255}}}
\newcommand{\bench}{\textbf{\gradientRGB{RefEdit-Bench}{0,76,153}{102,178,255}}}
\newcommand{\refeditsd}{\textbf{\gradientRGB{RefEdit-SD3}{0,76,153}{102,178,255}}}

\definecolor{cvprblue}{rgb}{0.86, 0.08, 0.24}

\usepackage[pagebackref,breaklinks,colorlinks,allcolors=cvprblue]{hyperref}
\usepackage{booktabs}       
\usepackage{amsfonts}       
\usepackage{nicefrac}       
\usepackage{microtype}      
\usepackage{xcolor}         
\usepackage{graphicx} 
\usepackage{amssymb} 
\usepackage{array} 
\usepackage{makecell} 
\usepackage{textcomp}  
\usepackage{changepage} 
\usepackage{scalerel}  

\usepackage{arydshln} %

\usepackage[normalem]{ulem}
\usepackage{multirow}

\def\paperID{} 
\def\confName{ICCV}
\def\confYear{2025}

\title{\refedit: A Benchmark and Method for Improving Instruction-based Image Editing Model on Referring Expressions}

\author{Bimsara Pathiraja\thanks{equal contribution} \quad Maitreya Patel\footnotemark[1] \quad Shivam Singh \quad Yezhou Yang \quad Chitta Baral\\
Arizona State University\\
{\tt\small \{bpathir1, maitreya.patel, ssing631, yz.yang, chitta\}@asu.edu} \\\\
\url{http://refedit.vercel.app}
}

\def\emojix{\scalerel*{\includegraphics{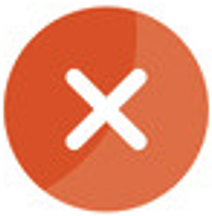}}{\LARGE\textrm{\textbigcircle}}}

\def\emojij{\scalerel*{\includegraphics{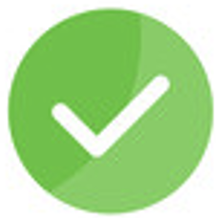}}{\LARGE\textrm{\textbigcircle}}}

\begin{document}


\twocolumn[{%
\renewcommand\twocolumn[1][]{#1}%
\maketitle

\centering
\captionsetup{type=figure}
\includegraphics[width=\linewidth]{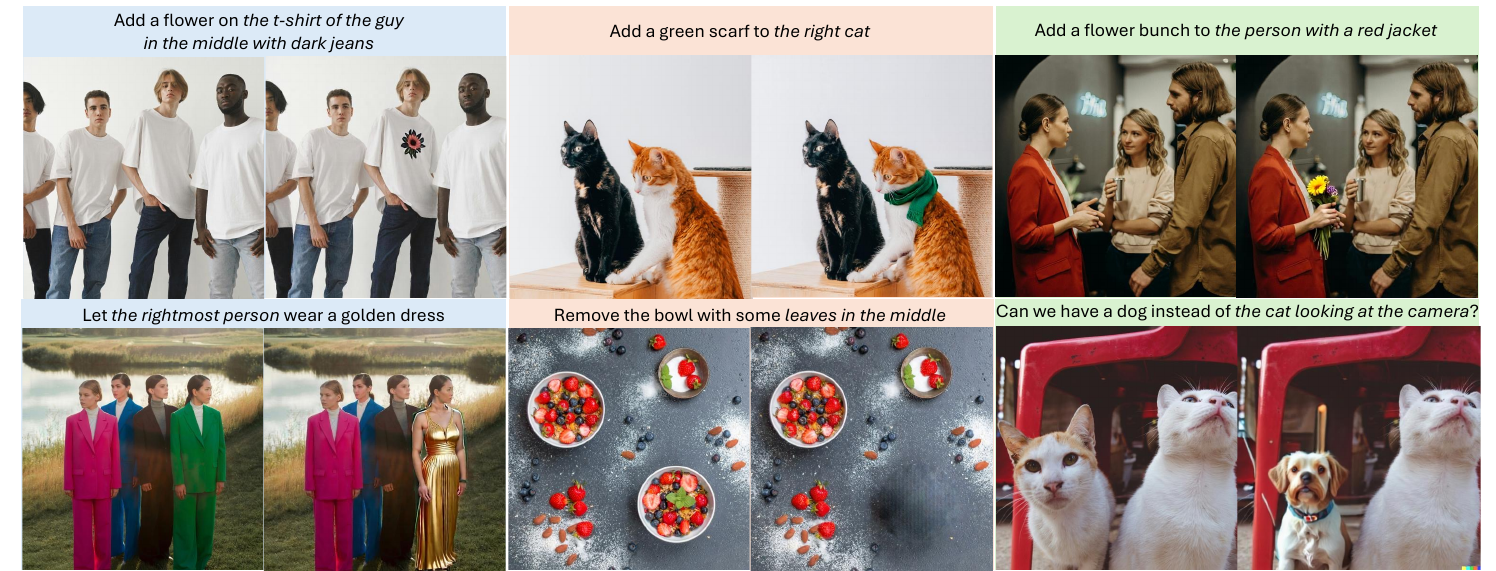}

\captionof{figure}{RefEdit is a referring expression-based image editing benchmark and a finetuned model. Our proposed RefEdit model can accurately identify the entity of interest and perform accurate edits.}
\label{fig:teaser}
\vspace{8pt}
}]

\renewcommand{\thefootnote}{\fnsymbol{footnote}} 
\footnotetext[1]{These authors contributed equally to this work.}

\begin{abstract}

Despite recent advances in inversion and instruction-based image editing, existing approaches primarily excel at editing single, prominent objects but significantly struggle when applied to complex scenes containing multiple entities. To quantify this gap, we first introduce \bench, a rigorous real-world benchmark rooted in RefCOCO, where even baselines trained on millions of samples perform poorly.
To overcome this limitation, we introduce \refedit\ -- an instruction-based editing model trained on our scalable synthetic data generation pipeline.
Our \refedit, trained on only 20,000 editing triplets, outperforms the Flux/SD3 model-based baselines trained on millions of data. 
Extensive evaluations across various benchmarks demonstrate that our model not only excels in referring expression tasks but also enhances performance on traditional benchmarks, achieving state-of-the-art results comparable to closed-source methods.
We release data \& checkpoint for reproducibility.

\end{abstract}    
\section{Introduction}
\label{sec:introduction}

The field of AI for content creation (AICC) stands at the forefront of technological innovation, propelled by unprecedented advances in visual generative models~\cite{rombach2022high, saharia2022photorealistic, esser2024scaling, patel2024eclipse, polyak2024movie, patel2024lambda, singer2023makeavideo}.
Techniques such as diffusion~\cite{dhariwal2021diffusion, song2020denoising} and flow-based models~\cite{liu2023flow, lipman2023flow, patel2024steeringrectifiedflowmodels} have revolutionized image editing, enabling precise semantic transformations guided by user queries. 
Early efforts centered on inversion-based strategies—like Prompt-to-Prompt~\cite{hertz2022prompt}, Null-Text Inversion~\cite{mokady2023null}, and edit-friendly DDPM-inversion \citep{ho2020denoising} which map reference image latents into noise and deterministically reconstruct the original. 
While these methods laid the critical groundwork, they are hindered by significant computational demands, prolonged processing times, and the need for extensive hyperparameter tuning. 
In response, instruction-based approaches, including InstructPix2Pix~\cite{brooks2023instructpix2pix}, UltraEdit~\cite{zhao2025ultraedit}, and OmniEdit~\cite{wei2024omniedit}, have gained prominence by fine-tuning diffusion models on millions of image editing triplets derived from inversion-based techniques, offering a more efficient alternative despite inheriting some foundational limitations.

These advancements, however, reveal a persistent weakness: 
current methods struggle to maintain precision in complex scenarios, particularly when images feature multiple similar entities (i.e., ``two cats''). 
Inversion-based strategies, though innovative, are computationally intensive and falter when tasked with isolating specific regions, a flaw that propagates to instruction-based models trained on their outputs. 
This imprecision often results in failed edits or unintended modifications spilling into adjacent areas. 
Synthetic data has been leveraged extensively across various domains beyond image editing~\cite{patel2024tripletclip, long2024llms}; however, existing pipelines within image editing rarely address multi-object complexity, further exacerbating their shortcomings. 
As a result, the field faces an unresolved bottleneck that limits the practical utility of image editing technologies in real-world applications.

We identify this bottleneck as the referring expression challenge—a problem well-explored in discriminative tasks like segmentation~\cite{yu2016modelingcontextreferringexpressions} but largely neglected in the generative domain of image editing. 
To confront this, we introduce \bench, a rigorous benchmark designed to evaluate image editing under both straightforward and demanding conditions. 
By repurposing the RefCOCO dataset~\cite{yu2016modelingcontextreferringexpressions}, we ensure real-world relevance, capturing intricate scenarios with multiple entities. 
As evidenced in Table~\ref{table:viescore}, leading baselines—such as MagicBrush \citep{zhang2024magicbrushmanuallyannotateddataset}, InstructDiffusion \citep{geng2023instructdiffusiongeneralistmodelinginterface} , and UltraEdit \citep{zhao2025ultraedit} — experience marked performance declines on our benchmark, highlighting its complexity. 

\begin{figure*}[!ht]
    \centering
    \includegraphics[width=1.0\linewidth]{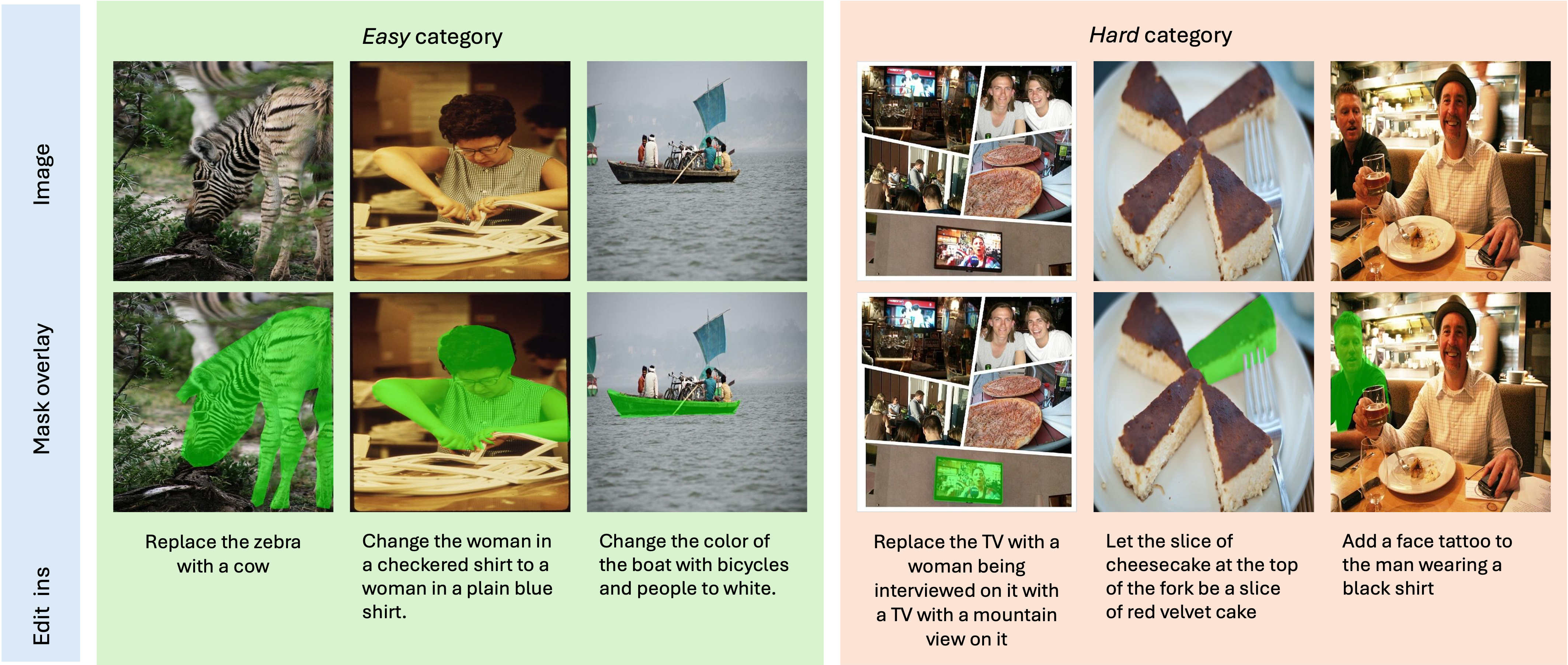}
    \caption{Three samples from each of \textit{Easy} and \textit{Hard} categories and the overlayed mask of the interested object are shown. 
    } 
    \label{fig:test-data}
\end{figure*}

To address this challenge, we propose a scalable synthetic data generation pipeline tailored to the referring expression challenge, harnessing GPT-4o \citep{achiam2023gpt} to craft sophisticated editing tasks and integrating Grounded Segment Anything \citep{ren2024grounded} with primarily a Flux-based inversion-free editing methodology (FlowChef \citep{patel2024steeringrectifiedflowmodels}) for precise, controlled edits. 
Using this pipeline, we fine-tune diffusion models—Stable Diffusion v1.5 \citep{rombach2022high} and Stable Diffusion 3 \citep{esser2024scaling}—to create \refedit, a model that redefines the state-of-the-art in image editing.
Extensive experiments, human evaluations, and detailed analyses confirm \refedit’s superiority, with our Stable Diffusion v1.5 variant outperforming baselines trained on vastly larger datasets and more advanced architectures like SDXL \citep{podell2023sdxl} and SD3. 
This breakthrough not only overcomes the referring expression challenge but also sets a new standard for efficiency and effectiveness in visual generative modeling. 
In summary, our contributions are:
\begin{itemize}
    \item We present \bench, a novel benchmark for realistic, referring expression-driven image editing.
    \item We introduce a novel, FlowChef and Inpaint Anything \citep{yu2023inpaint}-powered pipeline for automated, scalable synthetic data generation.
    \item Through comprehensive evaluation, we establish \refedit\ as the most human-preferred model, achieving exceptional success rates and outperforming existing methods.
\end{itemize}

\begin{table*}[t!]
  \centering
  \large
  \resizebox{\textwidth}{!}{
\begin{tabular}{ccccccccc}
  \\
  \toprule
  \Large{\bf Datasets} 
  & \makecell[c]{\Large{\bf Automatic} \\ \Large{\bf generated?}}  
  & \Large{\bf \#Edits} 
  & \makecell[c]{\Large{\bf \#Editing types}}
  & \makecell[c]{\Large{\bf Multiple similar} \\ \Large{\bf entities?}} 
  & \makecell[c]{\Large{\bf Avg. \#similar} \\ \Large{\bf entities}} 
  & \makecell[c]{\Large{\bf Use of} \\ \Large{\bf referring expressions?}} \\
  \midrule
  
  \makecell[c]{\Large{InstructPix2Pix~\citep{brooks2023instructpix2pixlearningfollowimage}}} 
  & \makecell[c]{\emojij} 
  & \makecell[c]{\Large{450K}} 
  & \makecell[c]{\Large{4}} 
  & \makecell[c]{\emojix} 
  & \makecell[c]{\Large{1}} 
  & \makecell[c]{\emojix} \\

  \makecell[c]{\Large{MagicBrush~\citep{zhang2024magicbrushmanuallyannotateddataset}}} 
  & \makecell[c]{\emojix} 
  & \makecell[c]{\Large{10K}} 
  & \makecell[c]{\Large{7}} 
  & \makecell[c]{\emojix} 
  & \makecell[c]{\Large{1}} 
  & \makecell[c]{\emojix} \\

  \makecell[c]{\Large{HQ-Edit~\citep{hui2024hq}}} 
  & \makecell[c]{\emojij} 
  & \makecell[c]{\Large{200K}} 
  & \makecell[c]{\Large{7}} 
  & \makecell[c]{\emojix} 
  & \makecell[c]{\Large{1}} 
  & \makecell[c]{\emojix} \\

  \makecell[c]{\Large{{UtraEdit ~\citep{zhao2025ultraedit}}}} 
  & \makecell[c]{\emojij} 
  & \makecell[c]{{\Large{4M}}} 
  & \makecell[c]{{\Large{9+}}} 
  & \makecell[c]{\emojix} 
  & \makecell[c]{\Large{1}} 
  & \makecell[c]{\emojix} \\

  \makecell[c]{\Large{{InstructDiffusion \citep{geng2023instructdiffusiongeneralistmodelinginterface}}}} 
  & \makecell[c]{\emojij} 
  & \makecell[c]{{\Large{425K}}} 
  & \makecell[c]{{\Large{6}}} 
  & \makecell[c]{\emojix} 
  & \makecell[c]{\Large{1}} 
  & \makecell[c]{\emojix} \\

  \makecell[c]{\Large{{HIVE \citep{zhang2024hive}}}} 
  & \makecell[c]{\emojij} 
  & \makecell[c]{{\Large{1.1M}}} 
  & \makecell[c]{{\Large{5+}}} 
  & \makecell[c]{\emojix} 
  & \makecell[c]{\Large{1}} 
  & \makecell[c]{\emojix} \\

  \makecell[c]{\Large{{OmniEdit \citep{wei2024omniedit}}}} 
  & \makecell[c]{\emojij} 
  & \makecell[c]{{\Large{1.2M}}} 
  & \makecell[c]{{\Large{7}}} 
  & \makecell[c]{\emojix} 
  & \makecell[c]{\Large{1}} 
  & \makecell[c]{\emojix} \\

  \midrule
  \makecell[c]{\Large{\textbf{RefEdit-Data}} (ours)} 
  & \makecell[c]{\textbf{\Large{\emojij}}} 
  & \makecell[c]{\textbf{\Large{20K}}} 
  & \makecell[c]{\textbf{\Large{5}}} 
  & \makecell[c]{\emojij} 
  & \makecell[c]{\Large{\textbf{3}}} 
  & \makecell[c]{\emojij} \\
  
\bottomrule
\end{tabular}

  }
\caption{Comparison of different image editing datasets.}
  
  \label{tab:compare_dataset}
\end{table*}
\section{Related works}
\label{sec:related_works}

\paragraph{Inversion-based editing.}

Inversion-based image editing leverages generative models to modify existing images by mapping them into a latent space, applying desired alterations, and then reconstructing the edited images. Prompt-to-Prompt \cite{hertz2022prompt} leverages cross attention maps to edit images by solely modifying the textual prompts. Negative-text Inversion \citep{miyake2023negative} and Edict \citep{wallace2023edict} uses DDIM/DDPM inversions to preserve the original image while editing the image. Other methods like LEdits \citep{tsaban2023ledits}, Sega \citep{brack2023sega}, Null-text inversion \citep{mokady2023null}, Direct Inversion \citep{ju2023direct} provides more efficient and reliable inversion based image editing methods. Still, the inversion based techniques suffer from leaking the editing to background areas and poor localization abilities. 
Notably we utilize FlowChef \citep{patel2024steeringrectifiedflowmodels} for our pipeline as it provides a gradient- and inversion-free framework for image editing.

\paragraph{Region-based controlled editing.}

Region-based editing approaches perform image modifications within user-provided masked regions guided by textual instructions. Blended Diffusion \citep{Avrahami_2022, Avrahami_2023} facilitates local inpainting based on user-supplied captions and masks. SDEdit \citep{meng2021sdedit} leverages pre-trained models to iteratively apply noise and denoising to real images guided by text prompts. DiffEdit \citep{couairon2022diffedit} integrates DDIM inversion for automatic mask generation to preserve image backgrounds during editing. MAGEdit \citep{mao2024mag} optimizes latent noise features via mask-based cross-attention constraints, progressively improving local alignment with desired textual prompts. Despite their effectiveness, region-based methods require additional user effort in mask annotation.

\paragraph{Instruction-based editing.}

Instruction-based editing methods allow users to modify images solely through textual descriptions of intended changes. InstructPix2Pix \citep{brooks2023instructpix2pixlearningfollowimage} utilizes image prompts and GPT-3-generated editing instructions \citep{brown2020languagemodelsfewshotlearners}, alongside image synthesis techniques from Prompt2Prompt \citep{hertz2022prompttopromptimageeditingcross}. Subsequently, MagicBrush \citep{zhang2024magicbrushmanuallyannotateddataset} introduced manually annotated datasets for InstructPix2Pix coupled with fine-tuned model checkpoints. InstructDiffusion \citep{geng2023instructdiffusiongeneralistmodelinginterface} extends beyond editing, incorporating broader computer vision tasks like segmentation and keypoint detection. OmniEdit \citep{wei2024omniedit} applies a specialist-to-generalist training strategy, consolidating diverse specialized editing capabilities into a single instruction-based model. UltraEdit \citep{zhao2025ultraedit} employs a large-scale, automatically generated dataset containing approximately 4 million samples for enhanced instruction-based image editing. Even though ReferDiffusion \citep{liu2024referring} addresses referring expressions based image editing, the training dataset is based on 400 initial images and its data generation pipeline is primarily manual, hindering scalability. Additionally, the work does not provide a benchmark for the community to facilitate comparative evaluations.

\section{Method}

This section introduces the \refedit\ synthetic data creation framework to train the \refedit\ model that can perform complex real-world edits. 
First, we will detail the preliminaries about diffusion-based instruction-guided image editing.
Next, we will present our synthetic data creation pipeline. 
Finally, we will share our training details. 

\subsection{Preliminaries}
Diffusion models are generative models that create images by reversing a diffusion process. In the forward process, Gaussian noise is incrementally added to an image across multiple time steps until it becomes pure noise. The reverse process, parameterized by a neural network, begins with noise and iteratively denoises it to produce a realistic image. The model is trained to predict the noise added at each step, minimizing the loss function:

\begin{equation}
     L = \mathbb{E}[\|\epsilon - \epsilon_\theta(z_t, t, c)\|^2],
\end{equation}

where $z_t$ represents the noisy latent at time $t$, $\epsilon$ is the actual noise, and $c$ denotes conditioning information, such as a text prompt.

\begin{figure}[t]
    \centering
    \includegraphics[width=1.0\linewidth]{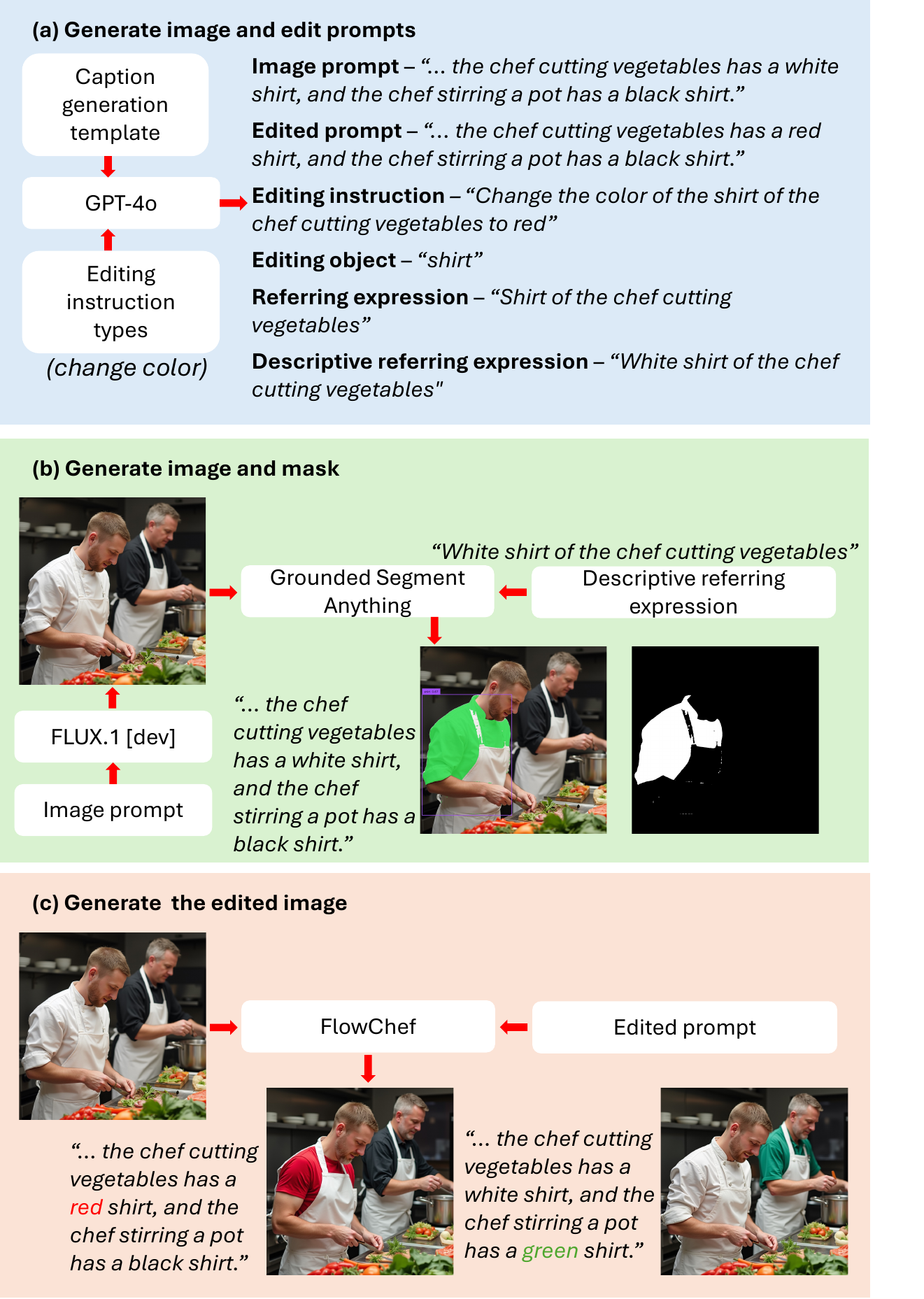}
    \caption{
    Overview of our pipeline for generating controlled image edits. GPT-4o \citep{openai2024gpt4ocard} produces the textual information. Based on that, FLUX \citep{flux2023} and Grounded Segment Anything \citep{ren2024grounded} identify masks based on these expressions. At last, we use  FlowChef~\cite{patel2024steeringrectifiedflowmodels}, mask-based inversion-free technique, for editing the images.
    }
    \label{fig:data-pipe}
\end{figure}

\begin{figure}[t]
    \centering
    \includegraphics[width=1.0\linewidth]{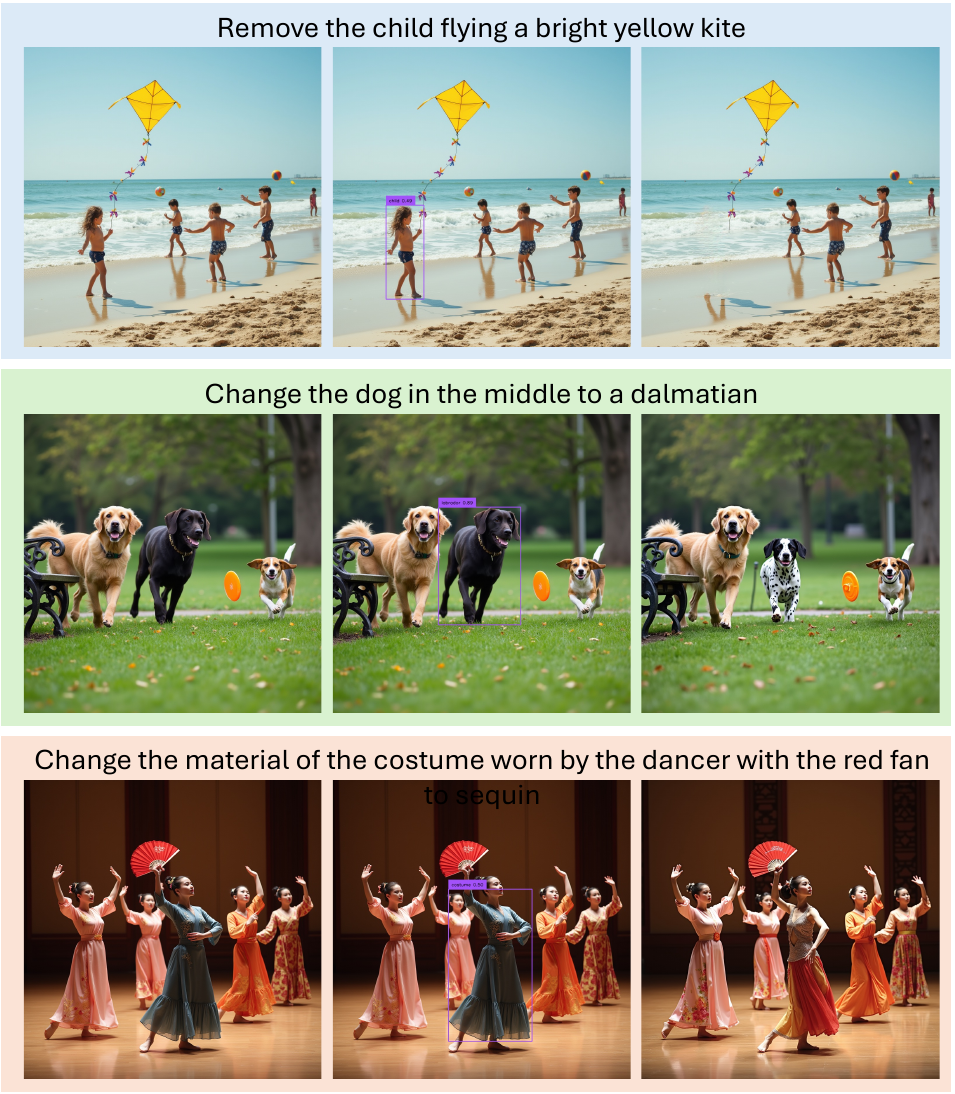}
    \caption{Two samples from our training dataset. From left to right: the initial image, the bounding box highlighting the object of interest, and the edited image. 
    }
    \label{fig:train-data}
\end{figure}

Instructpix2pix extends Stable Diffusion to enable text-guided image editing by conditioning the diffusion process on both an input image $c_I$ and a text instruction $c_T$. This modification allows the model to generate edited images that align with textual descriptions while retaining relevant aspects of the original image. The training objective is adapted to:

\begin{equation}
   L = \mathbb{E}_{\mathcal{E}(x),\mathcal{E}(c_I),c_T,\epsilon\sim\mathcal{N}(0,1),t}\left[\,\|\epsilon - \epsilon_\theta(z_t, t, \mathcal{E}(c_I), c_T)\|_2^2\,\right]
\end{equation}

where $\mathcal{E}(c_I)$ is the encoded input image. By leveraging a synthetic dataset comprising image pairs and corresponding edit instructions, Instructpix2pix learns to apply precise, instruction-driven modifications, offering an efficient and flexible approach to image editing without requiring per-example fine-tuning.

\subsection{RefEdit-Data: Synthetic data generation pipeline}

\begin{table*}
    \centering
    \begin{tabular}{p{2.5cm} p{4cm} p{6cm}}
       \toprule
        \textbf{Editing Tasks}        & \textbf{Definition}       & \textbf{Instruction example}      \\
        \midrule
          Change color & Describes the new color of an object instance. The initial color may or may not be given. & Can the horse which has a blue color rein be white? \newline Change the shirt color of the tennis player on right to red.\\
         \midrule
          Change object & Describes the object to be replaced with a new object in its place & Let the coffee drink be a matcha drink. \newline Replace the computer monitor with two green lines on it with a computer monitor displaying a vibrant beach scene.
         \\
         \midrule
         Add object & Describes a new object to add by the location the object should be added.
         & Put a red cap on the elderly man with a blue apron. \newline Add a bird to the bench next to the bench with the bird on it.
         \\
         \midrule
          Remove object & Describes the object instance that should be removed. 
         & Remove the half full mug of beer to the left of the pizza. \newline Get rid of the woman in a yellow shirt.
         \\
         \midrule
          Change texture& Describes how to change the texture, pattern or material of an object instance 
         & Turn the man wearing surfing suit and holding surfboard into a silver statue. \newline Let the man with the red t-shirt who is playing on his phone wear a stripped red shirt.
         \\
         \bottomrule
    \end{tabular}
    \caption{Task Definitions and examples from RefEdit-Bench.}
    \label{tab:task_definitions}
\end{table*}

\begin{figure}[t]
    \centering
    \includegraphics[width=\linewidth]{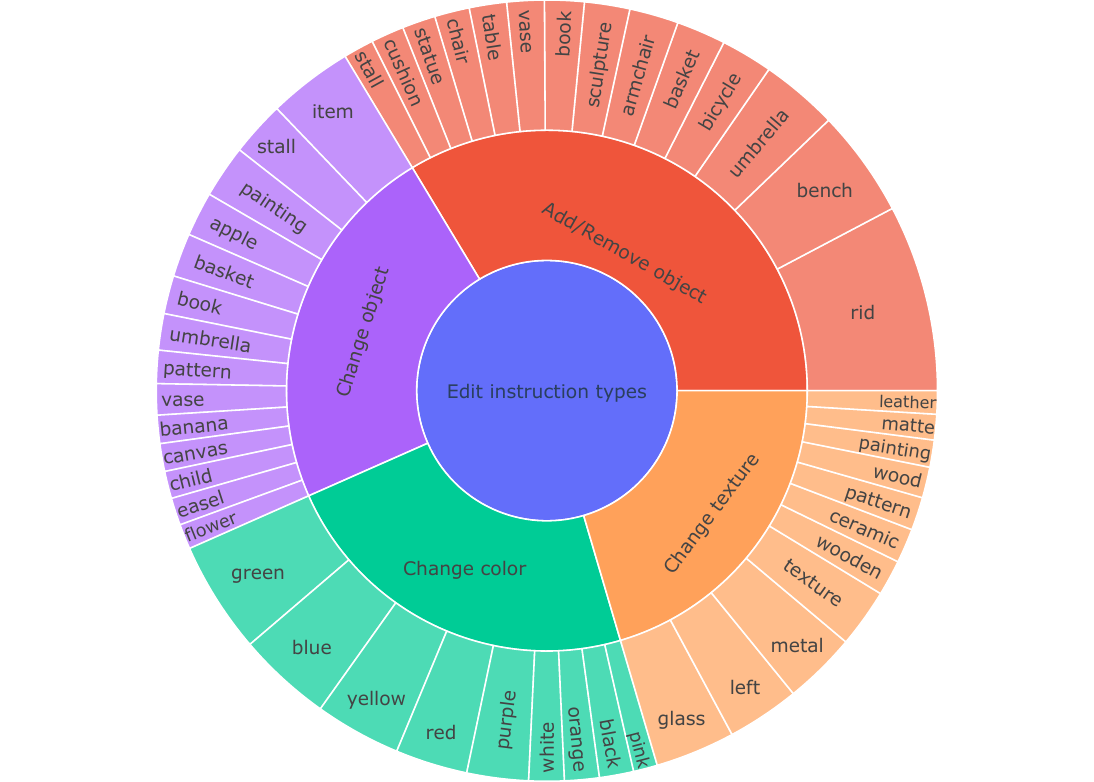}
    \caption{An overview of keywords in edit instructions. The inner circle depicts the types of edits and outer circle showcases the most frequent words used within each type.}
    \label{fig:edit_ins_sun}
\end{figure}

The central challenge of instruction-based image editing is the data.
Therefore, existing works rely on some form of synthetic data creation (see Table~\ref{tab:compare_dataset}).
One way to create such editing triplets is inversion-based editing methods, which do not require any training. 
However, inference-time editing is a very ill-posed problem statement. 
It requires careful hyperparameter selection for each prompt, and inversion is not guaranteed to work. 
Despite these challenges, existing approaches utilize these methods to create million-scale diverse editing triplets.
Furthermore, these inversion-based editing frameworks fail measurably in referring expressions (see Table~\ref{table:viescore}).
Hence, the challenges of inversion-based editing methods limit instruction-based editing and require a lot of careful data creation process.
Additionally, we found that existing referring expression datasets (e.g., RefMSCOCO) were very difficult for the existing method, and the failure rate was very high, which made them useless.
Therefore, we propose a new and simplified data generation framework.

\begin{table*}[ht]
    \centering
    \begin{tabular}{l |ccc|ccc}
        \toprule
        \multirow{2}{*}{\textbf{Method}} & \multicolumn{3}{c|}{\textbf{Easy}} & \multicolumn{3}{c}{\textbf{Hard}} \\
        & \textbf{SC\textsubscript{avg} $\uparrow$} & \textbf{PQ\textsubscript{avg} $\uparrow$} & \textbf{O\textsubscript{avg} $\uparrow$} & \textbf{SC\textsubscript{avg} $\uparrow$} & \textbf{PQ\textsubscript{avg} $\uparrow$} & \textbf{O\textsubscript{avg} $\uparrow$} \\
        \midrule
        \textbf{P2P \cite{hertz2022prompt}} & 2.7 & 5.46 & 2.31 & 1.71 & 5.3 & 1.66 \\
        \textbf{P2P (Direct Inversion)} & 2.96 & 5.84 & 2.65 & 2.42 & 5.9 & 2.3 \\
        \textbf{PnP} & 4.25 & 5.59 & 3.55 & 2.45 & 5.62 & 2.27 \\
        \midrule
        \textbf{InstructPix2Pix} & 3.52 & 5.66 & 2.82 & 3.80 & 6.02 & 3.22 \\
        \textbf{MagicBrush} & 4.18 & \textbf{6.10} & 3.67 & 4.11 & 6.16 & 3.56 \\
        \textbf{HIVE} & 3.15 & 5.26 & 2.42 & 3.22 & 5.52 & 2.47 \\
        \textbf{InstructDiffusion} & 4.16 & 4.71 & 3.31 & 4.09 & 4.94 & 3.16 \\
        \textbf{HQ-Edit} & 2.58 & 5.39 & 2.37 & 2.95 & 5.21 & 2.6 \\
        \textbf{OmniEdit\textsuperscript{*}} & 4.18 & 6.15 & 3.68 & 4.07 & \underline{6.28} & 3.53 \\
        \textbf{InstructDiffusion-HA} & \underline{5.01} & \underline{5.88} & \underline{4.21} & \textbf{4.56} & 6.1 & \underline{3.75} \\
        \hdashline
        \textbf{\refedit} \textit{(ours)} & \textbf{5.47} & 5.85 & \textbf{4.68} & \underline{4.51} & \textbf{6.48} & \textbf{3.93} \\
        \midrule
        \textbf{CosXLEdit} & 3.2 & \underline{6.73} & 2.88 & \underline{3.44} & \textbf{6.65} & \underline{2.88} \\
        \textbf{FLUX-OmniEdit} & 1.95 & 2.46 & 1.31 & 1.55 & 1.92 & 0.89 \\
        \textbf{UltraEdit} & \underline{3.68} & \textbf{6.98} & \underline{3.4} & 2.93 & \underline{6.23} & 2.59 \\
        \hdashline
        \textbf{\refeditsd} \textit{(ours)} & \textbf{5.46} & 6.36 & \textbf{4.99} & \textbf{4.47} & 6.21 & \textbf{3.98} \\
        \bottomrule
    \end{tabular}
    \caption{Modified VIEScore evaluation results on RefEdit benchmark for both \textit{Easy} and \textit{Hard} categories. The best value is bolded and the second-best value is underlined. \textbf{O\textsubscript{avg}} is the overall VIEScore. GPT-4o is used as the MLLM.}
    \label{table:viescore}
    
\end{table*}

In Figure~\ref{fig:data-pipe}, we outline the four key steps of our framework. 
For training data generation we start with generating all the text data. 
We utilize GPT-4o \cite{openai2024gpt4ocard} to generate the image prompt, edited prompt editing instruction, edited object, referring expression, and a more expanded version of the same referring expression. 
We group the image editing task into five different categories, as detailed in Table~\ref{tab:task_definitions}. 
Importantly, each category focuses on the referring expression.
We usually generate two sets of editing information per the same image prompt. 
In the template, we specify how to generate the image prompts, such as if there are two or more instances of the same object but with some differences. 
The difference could be due to the color (``A white shirt" or a ``black shirt"), a separate item (``person holding a knife" or ``a person holding a bag"), or an activity like in this case (``chef cutting vegetables" or ``chef stirring the pot"). 
We generate a 1024 \(\times\) 1024 image using FLUX \citep{flux2023}. 
The referring expression is used to generate the editing instruction, but we ask the LLM to generate a descriptive referring expression for ease of mask generation. 
Grounded SAM \citep{ren2024grounded} uses Segment Anything \citep{kirillov2023segany} and Grounding DINO \citep{liu2023grounding} to generate accurate masks of objects using referring expressions. 
The initial image, the editing instructions, and the mask are used for generating the edited image. We utilized Inpaint Anything \citep{yu2023inpaintanythingsegmentmeets} for object removals and FlowChef \citep{patel2024steeringrectifiedflowmodels} for other editing tasks. Adding object data were created by swapping the initial image and the edited image.
With the help of this pipeline, we create a 20000+ paired training dataset. 
Figure~\ref{fig:train-data} shares the three examples of our training dataset. 
Each image contains multiple similar entities that require some form of referring expression to pinpoint. 
Figure~\ref{fig:edit_ins_sun} shows that our \refedit\ data is very diverse and well distributed across the editing categories.

\subsection{Training Details}

We use the finetuning configuration of MagicBrush \citep{zhang2024magicbrushmanuallyannotateddataset} where the training starts with the pre-trained InstructPix2Pix model.
To avoid overfitting on our RefEdit-Data and to ensure that our method can generalize to other tasks, we combine MagicBrush data with our own generated data during the training. 
This results in the 30K amount of combined training data.
Specifically, we train the \refedit\ Stable Diffusion v1.5 variant on this data following the Eq. (1) for 24 epochs on 2 \(\times\) 80 GB NVIDIA A100 GPUs. 
Similarly, we train \refeditsd\ (UltraEdit) for 6000 iterations on 8 $\times$ 80 GB NVIDIA A100 GPUs.

\section{RefEdit-Bench}

To evaluate the models' performance on more complex and semantically rich image editing tasks, we introduce the \refedit\ Benchmark, a novel evaluation benchmark dataset designed to assess the capability of models in handling referring expression-based editing instructions. The benchmark leverages images sourced from the RefCOCO dataset \citep{yu2016modelingcontextreferringexpressions}, which provides a rich collection of images annotated with referring expressions. Each image in \refedit\ is carefully selected to ensure diversity in scenes, objects, and contexts. Additionally, we manually craft detailed and varied editing instructions for each image, encompassing tasks such as changing color, changing objects, adding objects, removing objects, and changing texture. 

The benchmark is divided into \textit{Easy} and \textit{Hard} categories, in which each category has 100 images. 
In the \textit{Easy} category, the images had primarily one object. 
Even if there were multiple instances of the same object, mostly the interested object occupies a significant portion of the image or can easily be identified. 
In the \textit{Hard} category, there are multiple instances of the same object, and mostly those instances occupy the same areas of the image, making isolating the correct instance hard for the model. 
Some chosen examples from both \textit{Easy} and \textit{Hard} are shown in Figure \ref{fig:test-data}.

\section{Experiments}

In this section, we first discuss the baselines and evaluation setup. Later, we make extensive comparisons on \bench\ and PIE-Bench \citep{ju2023direct} along with human evaluations.

\subsection{Experimental setup}

\begin{table}[h]
    \centering
    \resizebox{\linewidth}{!}{%
        \begin{tabular}{l | ccc}
            \toprule
            \textbf{Model} & \textbf{SC\textsubscript{avg} $\uparrow$} & \textbf{PQ\textsubscript{avg} $\uparrow$} & \textbf{O\textsubscript{avg} $\uparrow$} \\
            \midrule
            \textbf{InstructPix2Pix} & 4.00 & \underline{6.62} & 3.96 \\
            \textbf{MagicBrush} & \underline{5.02} & 6.80 & 4.79 \\
            \textbf{InstructDiffusion} & 4.49 & 5.51 & 4.06 \\
            \textbf{OmniEdit} & 5.18 & \textbf{7.22} & \underline{5.12} \\
            \textbf{InstructDiffusion-HA} & 4.69 & 6.86 & 4.54 \\ 
            \hdashline
            \textbf{\refedit} \textit{(ours)} & \textbf{5.65} & 6.46 & \textbf{5.21} \\
            \midrule
            \textbf{CosXLEdit} & \underline{5.12} & \textbf{7.78} & \underline{5.11} \\
            \textbf{FLUX-OmniEdit} & 2.37 & 3.59 & 2.14 \\
            \hdashline
            \textbf{\refeditsd} \textit{(ours)} & \textbf{6.04} & \underline{6.47} & \textbf{5.70} \\
            \bottomrule
        \end{tabular}
    }
    \caption{VIEScore evaluation results on PIE-bench. The best value is bolded and the second-best value is underlined. GPT-4o is used as the MLLM.}
    \label{table:pie}

\end{table}

\begin{figure*}[t]
    \centering
    \includegraphics[width=1.0\linewidth]{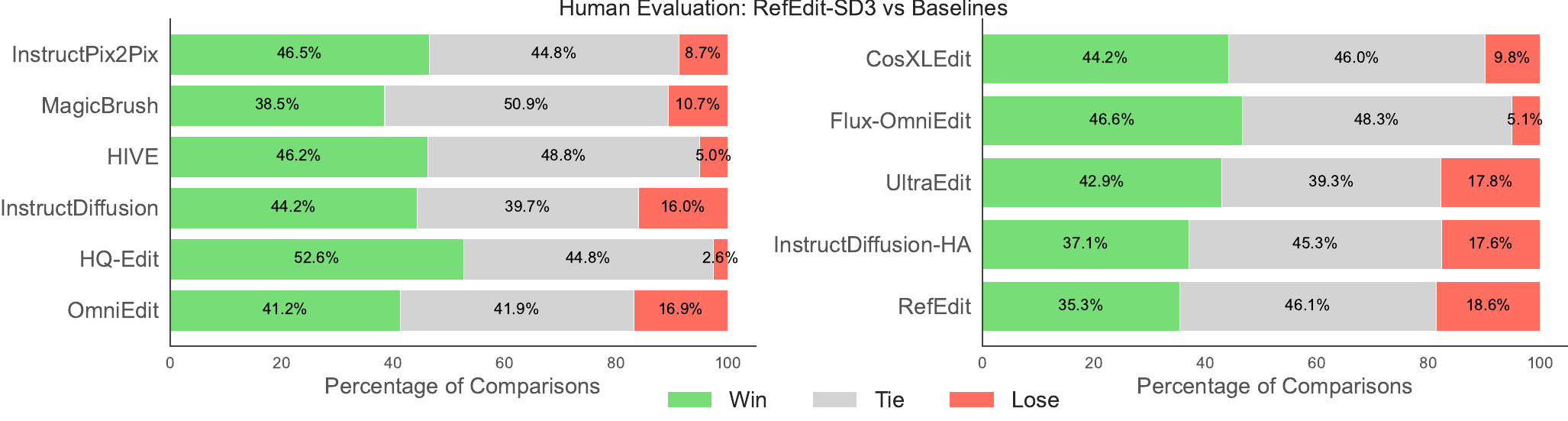}
        \caption{Human preference analysis for image editing, \refeditsd\ vs. all. Here, win indicates the winning rate for \refeditsd.}
    \label{fig:human-eval}
\end{figure*}

\paragraph{Baselines.} 
We evaluate multiple instruction-based image editing models across the various base models at scale. 
We select pre-trained models built on top of Stable Diffusion v1.5, such as InstructPix2Pix, MagicBrush, InstructDiffusion, HQ-Edit, and HIVE.
Most of these baselines use the InstructPix2Pix as the base model and finetune it using respective proposed data.
Similarly, we train additional baseline OmniEdit on SDv1.5 using the released 1.2M editing dataset. 
Apart from this, we consider bigger baselines such as CosXLEdit (Stable Diffusion XL), UltraEdit (Stable Diffusion 3), and Flux-OmniEdit.
As OmniEdit model is not public, we utilize the open-source replication\footnote{\href{https://huggingface.co/sayakpaul/FLUX.1-dev-edit-v0}{https://huggingface.co/sayakpaul/FLUX.1-dev-edit-v0}}.
We present two variants, \refedit\ \& \refeditsd, trained on our proposed synthetic data.

\paragraph{Evaluation Metrics.}
We perform extensive evaluations on our proposed \bench\ and PIE-Bench for referring expression and general tasks, respectively.
We incorporate PIE-Bench as it is, which utilizes several different metrics focusing on background preservation and image editing.
Importantly, it utilizes the ground truth mask to determine the region of interest.
However, as CLIP is not good at referring expression, we instead utilize VIEScore \citep{ku2024viescoreexplainablemetricsconditional} as an alternative evaluation metric for both benchmarks, which is a training-free visual instruction-guided metric leveraging multimodal LLMs (MLLMs).
It contains two metrics: Semantic Consistency ($SC$) and Perceptual Quality ($PQ$). $SC$ determines the alignment of the edited image with the editing instruction and $PQ$ determines the authenticity and the naturalness of the image. Following prompting templates from OmniEdit \citep{wei2024omniedit}, we prompt GPT-4o \citep{openai2024gpt4ocard} on $SC$ and $PQ$. And we report the average performance. 
At last, we calculate the overall score as: $O = \sqrt{SC \times PQ}$.
To further improve the referring expression ability of GPT-4o, after PIE-Bench, we incorporate the ground truth mask to extract the region of interest.
We call this metric Modified VIEScore.

\subsection{Quantitative Evaluation}

\paragraph{\bench\ Evaluations.}
We provide the VIEScore based evaluations in Table~\ref{table:viescore} for easy and hard categories.
It can be observed that the \refedit\ (SDv1.5) variant consistently outperforms the baselines by significant margins on both categories.
Interestingly, MagicBrush trained on small amount of human annotated data performs similar to OmniEdit's 1.2 million synthetic data. 
While \refedit\ consistently tops the benchmark.
This shows the importance of small but high-quality datasets.
Interestingly, bigger models (CosXLEdit, UltraEdit, etc.) perform subpar even worse than the SDv1.5 based methods and observes the biggest drop.
We attribute this behavior to potential overfitting. 
As existing training datasets do not contain complex image editing often requiring referring expression, larger models tend to overfit. 
Importantly, adding only 20000 of our synthetic data improves the performance over UltraEdit significantly across both categories. 

\begin{figure*}[ht]
    \centering
    \includegraphics[width=1.0\linewidth]{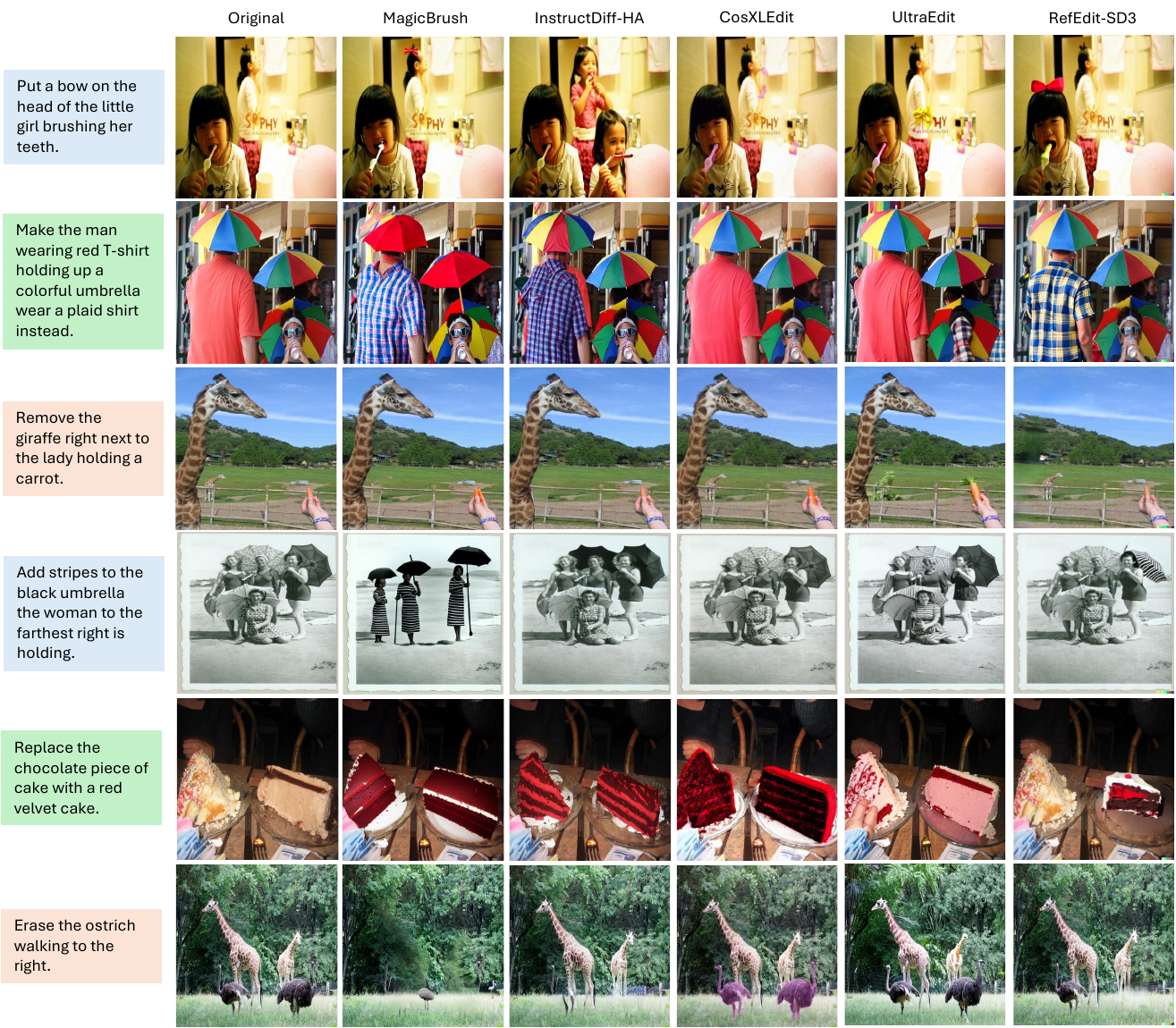}
    \caption{Qualitative results on image editing. The top 3 samples are from the \textit{Easy} category and the bottom 3 samples are from the \textit{Hard} category. As illustrated, our method attains the SOTA performance on comparison of all the methods.}
    \label{fig:qualitative_main}
\end{figure*}

\paragraph{PIE-Bench Evaluations.}
Apart from \bench, we evaluate the methods on standard image editing tasks using PIE-Bench (see Table~\ref{table:pie}).
It can be observed that \refedit\ further improves the performance over the baselines despite not containing any additional training data.
Importantly, OmniEdit achieves a 5.12 score with the help of a 1.2M dataset while \refedit\ gets 5.21 with only 20000 additional training data.
Our \refeditsd\ variant achieves the SOTA performance.
Importantly, all methods drop performance on \bench\ compared to the PIE-Bench.
That shows the difficulty of our \bench\ tasks. 

\subsection{Qualitative Evaluations.}
We perform human evaluations, where we ask the human annotators to pick the model that performs the task accurately or whether both models at hand lose or win.
Specifically,  randomly selected 400 pairs from \bench\ and asked the annotators to conduct A/B testing.
Figure~\ref{fig:human-eval} shows the human evaluation results.
It can be observed that our \refeditsd\ model significantly outperforms all baselines and is preferred consistently.
Furthermore, Figure~\ref{fig:qualitative_main} shows the qualitative results from \bench\ in both \textit{Easy} and \textit{Hard} categories.
While baselines either leak the edit to background regions or don't perform edits at all, our model \refeditsd\ can recognize the object of interest and consistently performs edits while maintaining the background information. We provide more examples in the appendix.

\section{Conclusion}

In this paper, we introduce \bench, a challenging benchmark designed specifically for evaluating image editing models on referring expressions, and \refedit, a novel model that significantly surpasses the existing state-of-the-art in this domain. By leveraging a carefully designed synthetic data generation pipeline, we demonstrate that training \refedit\ on a relatively small yet highly specialized dataset effectively outperforms existing baselines trained on considerably larger datasets. Our results underscore the potential of targeted, quality-focused synthetic data for improving model precision in complex editing scenarios. 
Future research could extend our pipeline to additional editing scenarios and explore further integration with multimodal AI frameworks.

\section*{Acknowledgment}
This work was supported by NSF RI grant \#2132724. We thank the NSF NAIRR initiative, Research Computing (RC) at Arizona State University (ASU), and cr8dl.ai for their generous support in providing computing resources. The views and opinions of the authors expressed herein do not necessarily state or reflect those of the funding agencies and employers.

{
    \small
    \bibliographystyle{ieeenat_fullname}
    \bibliography{main}
}



\begin{figure*}[t]
    \centering
    \includegraphics[width=1.0\linewidth]{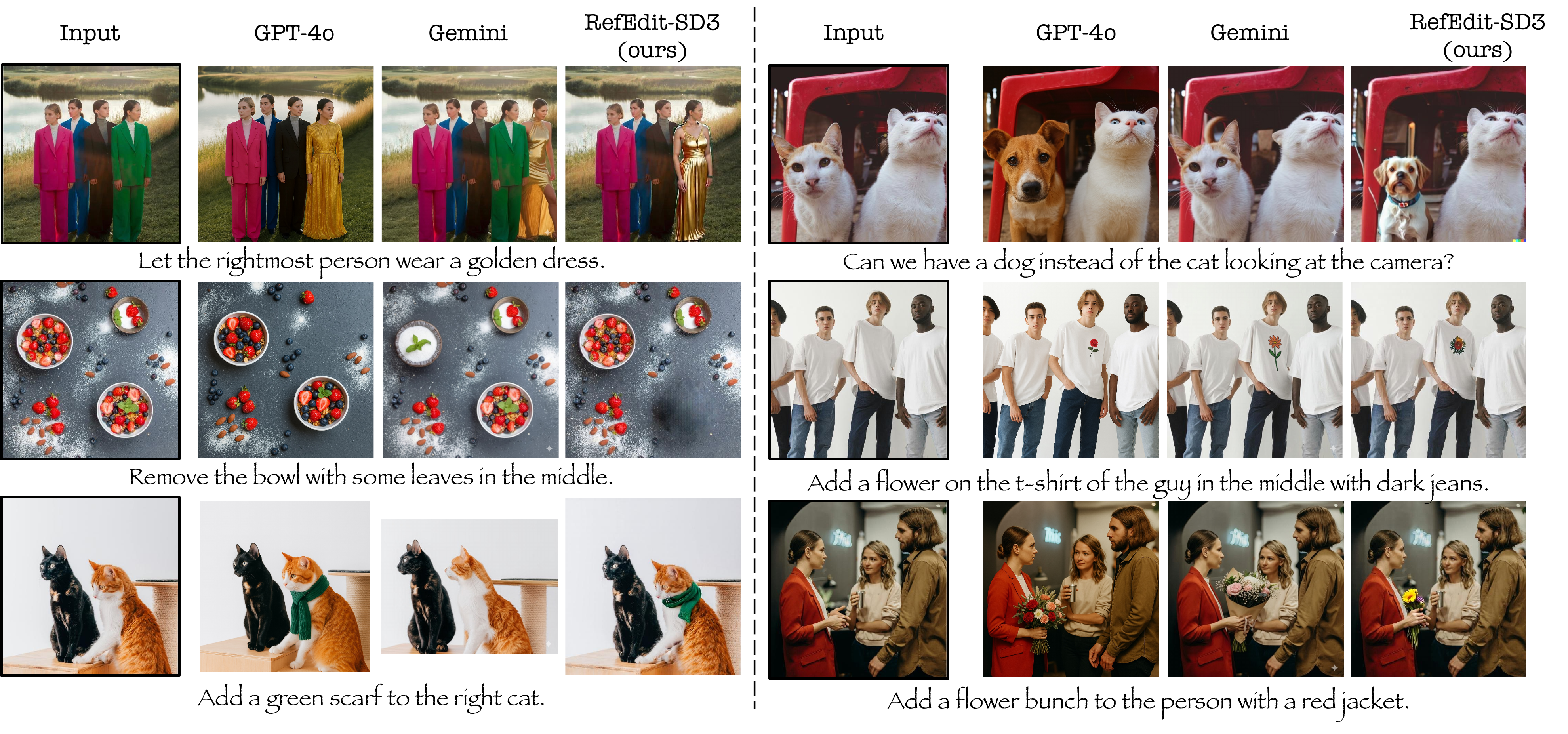}
    \caption{Qualitative comparisons with closed-source methods.}
    \label{fig:sc_mask_sup1}
\end{figure*}

\begin{figure*}[t]
    \centering
    \includegraphics[width=1.0\linewidth]{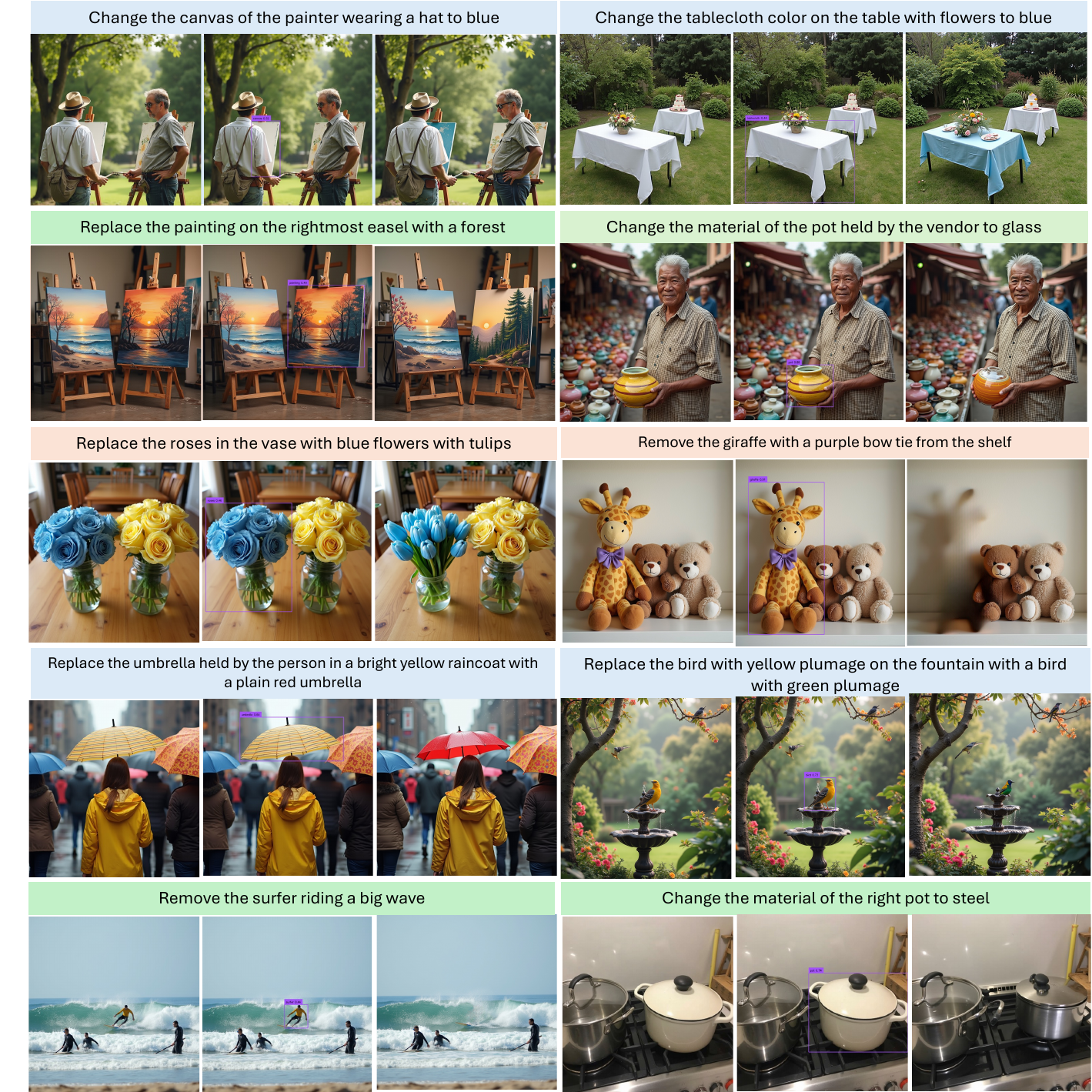}
    \caption{Additional training samples.}
    \label{fig:sc_mask_sup2}
\end{figure*}

\begin{table*}[ht]
    \centering
    \caption{Evaluation results on RefEdit benchmark for both \textit{Easy} and \textit{Hard} categories. The best value is bolded and the second-best value is underlined. }
    \label{table:pnp}
    \begin{tabular}{ll ccccccc}
        \toprule
        & & \textbf{Structure} & \multicolumn{4}{c}{\textbf{Background preservation}} & \multicolumn{2}{c}{\textbf{CLIP similarity}} \\
        \cmidrule(lr){3-3} \cmidrule(lr){4-7} \cmidrule(lr){8-9}
        & \textbf{Model} & \textbf{Distance $\downarrow$} & \textbf{PSNR $\uparrow$} & \textbf{LPIPS $\downarrow$} & \textbf{MSE $\downarrow$} & \textbf{SSIM $\uparrow$} & \textbf{Whole $\uparrow$} & \textbf{Edited $\uparrow$} \\
        \midrule
        
        \multirow{13}{*}{\rotatebox[origin=c]{90}{\textbf{Easy}}} 
        & InstructPix2Pix & 0.0305 & 21.40 & 0.1190 & 0.0129 & 0.7577 & 24.41 & 20.90 \\
        & MagicBrush & 0.0207 & 24.39 & 0.0701 & 0.0076 & 0.8065 & 25.39 & 20.94 \\
        & HIVE & 0.0287 & 22.23 & 0.1169 & 0.0104 & 0.7485 & 23.75 & 20.68 \\
        & InstructDiffusion & 0.0400	& 22.76	& 0.0900	& 0.0200	& 0.7800	& 24.34	& 20.39 \\
        & HQ-Edit & 0.1130 & 12.03 & 0.3418 & 0.0696 & 0.4913 & 20.48 & 18.33 \\
        & OmniEdit\textsuperscript{*} & 0.0190 & 24.80 & 0.0645 & 0.0070 & 0.8116 & 25.15 & 20.92 \\
        & InstructDiffusion-HA & 0.0252 &	24.95 & \underline{0.0598} & 0.0068 & 0.8143 & 24.73 & 20.85 \\
        & CosXLEdit & \underline{0.0137} & \textbf{26.60} & 0.0695 & \underline{0.0062} & \textbf{0.8962} & 25.21 & 20.79 \\
        & FLUX-Omni-Edit & 0.0400 &	20.48 & 0.1300 & 0.0200	& 0.7800 & 21.44 & 17.5 \\
    & UltraEdit & \textbf{0.0120} & 26.23 & 0.0740 & \textbf{0.0042} & 0.8358 & 25.29 & \underline{20.96} \\
    \hdashline
        & \refedit & 0.0199 & 24.81 & 0.0599 & 0.0064 & 0.8145 & \underline{25.48} & \textbf{21.07} \\
    & \refeditsd & 0.0239 & \underline{26.49} & \textbf{0.0572} & 0.0069 & \underline{0.8902} & \textbf{25.79} & 20.84 \\
        
        \midrule
        
        \multirow{13}{*}{\rotatebox[origin=c]{90}{\textbf{Hard}}} 
        & InstructPix2Pix & 0.0435 & 18.87 & 0.1664 & 0.0231 & 0.6775 & 25.60 & 19.97 \\
        & MagicBrush & 0.0274 & 20.56 & 0.1074 & 0.0151 & 0.7337 & 26.59 & \underline{20.21} \\
        & HIVE & 0.0367 & 20.01 & 0.1601 & 0.0173 & 0.6781 & 24.88 & 20.03 \\
        & InstructDiffusion & 0.0400 & 18.96 & 0.1300 & 0.0300 & 0.7000 & 25.62 & 19.36 \\
        & HQ-Edit & 0.1502 & 10.96 & 0.4127 & 0.0883 & 0.3789 & 20.88 & 17.8 \\
        & OmniEdit\textsuperscript{*} & 0.0248 & 20.80 & 0.1005 & 0.0140 & 0.7413 & 26.54 & 20.18 \\
        & InstructDiffusion-HA & 0.0226 & 21.12 & \underline{0.0886} & \underline{0.0128} & 0.7495 & 26.36 & 19.60 \\
        & CosXLEdit & 0.0267 & 21.61 & 0.1237 & 0.0240 & \underline{0.8241} & 26.65 & 19.91 \\
        & FLUX-OmniEdit & 0.0500	& 16.97 &	0.2100	& 0.0300	& 0.6700 & 21.02 & 16.05 \\
    & UltraEdit & \textbf{0.0144} & \textbf{23.64} & 0.1006 & \textbf{0.0067} & 0.7743 & \textbf{27.03} & 19.82 \\
    \hdashline
        & \refedit & \underline{0.0206} & 21.56 & \textbf{0.0868} & 0.0131 & 0.7531 & \underline{26.74} & \textbf{20.30} \\
    & \refeditsd & 0.0259 & \underline{22.15} & 0.0911 & 0.0152 & \textbf{0.8460} & 26.46 & 19.66 \\
        \bottomrule
    \end{tabular}
\end{table*}

\begin{figure*}[t]
    \centering
    \includegraphics[width=1.0\linewidth]{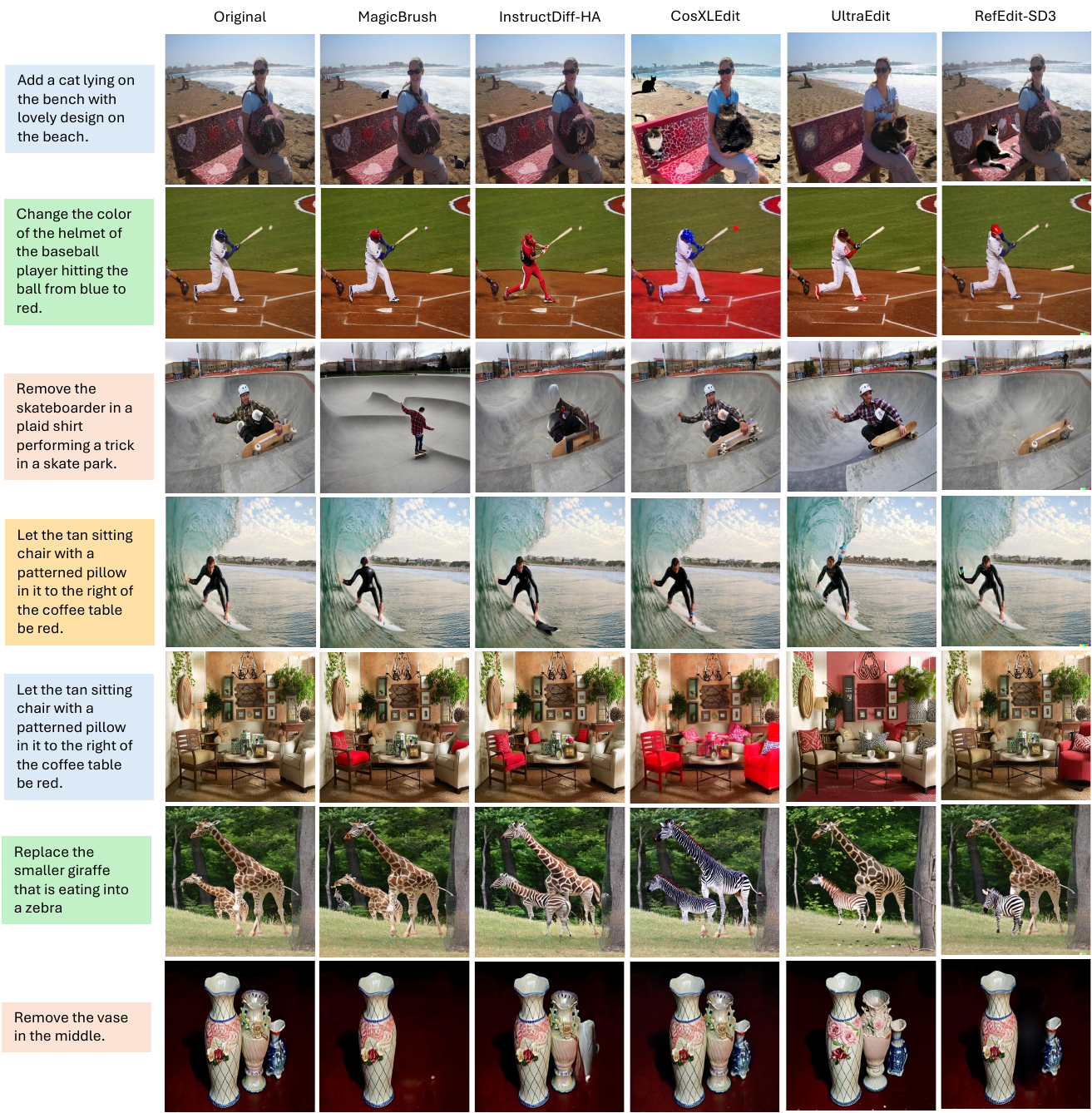}
    \caption{Qualitative results on image editing. The top 4 samples are from the \textit{Easy} category and the bottom 3 samples are from the \textit{Hard} category. As illustrated, our method attains the SOTA performance on comparison of all the methods.}
    \label{fig:qualitative_sup}
\end{figure*}


\begin{figure*}[t]
    \centering
    \includegraphics[width=1.0\linewidth]{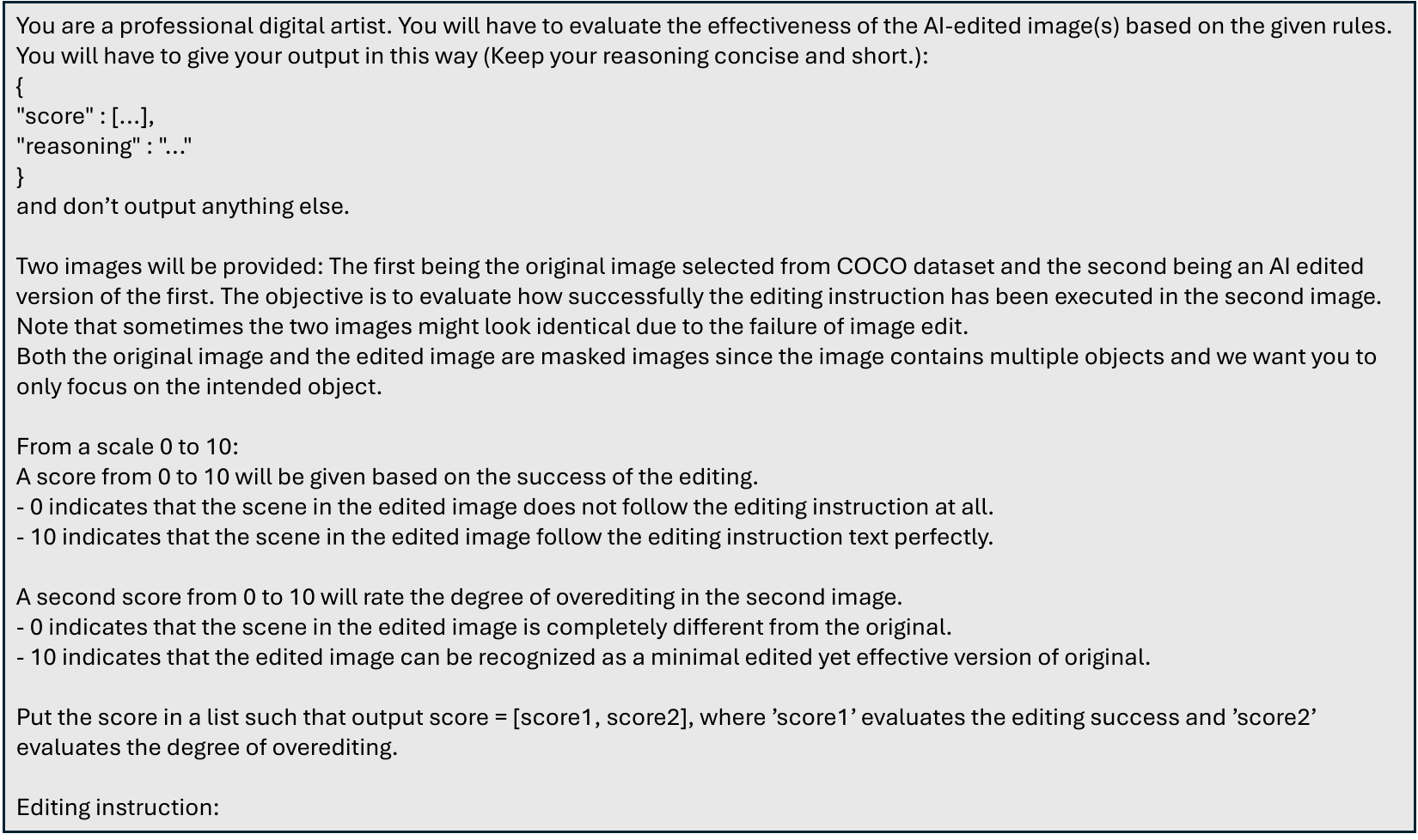}
    \caption{SC score prompt masked version.}
    \label{fig:sc_mask_sup3}
\end{figure*}


\begin{figure*}[t]
    \centering
    \includegraphics[width=1.0\linewidth]{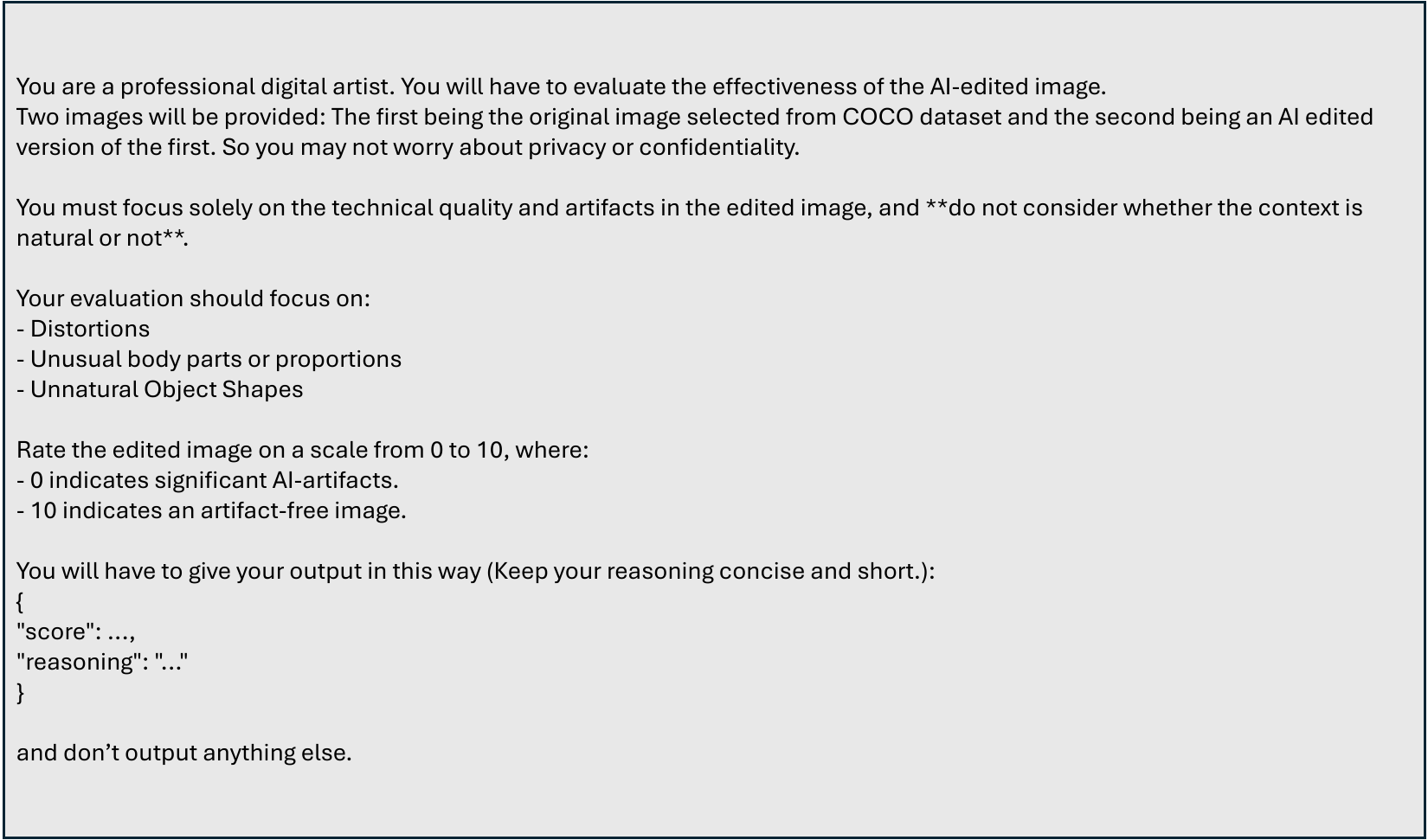}
    \caption{PQ score version.}
    \label{fig:pq_sup}
\end{figure*}

\end{document}